\documentclass[runningheads]{llncs}
\usepackage[table]{xcolor}
\usepackage{graphicx}

\usepackage{tikz}
\usepackage{comment}
\usepackage{amsmath,amssymb} 
\usepackage{color}
\usepackage{multirow}
\usepackage{wrapfig}
\usepackage[colorlinks,
            linkcolor=red, 
            anchorcolor=blue,  
            citecolor=green,       
            ]{hyperref}
\usepackage{subfigure}

\usepackage[accsupp]{axessibility}  

\begin{document}
\pagestyle{headings}
\mainmatter
\def\ECCVSubNumber{5032}  

\title{Instance As Identity: A Generic Online Paradigm for Video Instance Segmentation} 

\titlerunning{IAI: Instance As Identity for online VIS}
\author{Feng Zhu\inst{1,2,3}\thanks{Work done during an internship at Baidu.} \and
Zongxin Yang\inst{4} \and
Xin Yu \inst{3} \and
Yi Yang \inst{4} \and
Yunchao Wei\inst{5, 6}}
\authorrunning{F. Zhu et al.}
%
\institute{Baidu Research \and
ReLER, Centre for Artificial Intelligence, University of Technology Sydney \and
Australian Artificial Intelligence Institute, University of Technology Sydney \and
CCAI, College of Computer Science and Technology, Zhejiang University \and
Institute of Information Science, Beijing Jiaotong University \and
Beijing Key Laboratory of Advanced Information Science and Network\\
\email{Feng.Zhu@student.uts.edu.au}}
\maketitle

\begin{abstract}
Modeling temporal information for both detection and tracking in a unified framework has been proved a promising solution to video instance segmentation (VIS).
However, how to effectively incorporate the temporal information into an online model remains an open problem.  
In this work, we propose a new online VIS paradigm named Instance As Identity (IAI), which models temporal information for both detection and tracking in an efficient way. In detail, IAI employs a novel identification module to predict identification number for tracking instances explicitly. For passing temporal information cross frame, IAI utilizes an association module which combines current features and past embeddings. Notably, IAI can be integrated with different image models. We conduct extensive experiments on three VIS benchmarks. IAI outperforms all the online competitors on YouTube-VIS-2019 (ResNet-101 43.7 mAP) and YouTube-VIS-2021 (ResNet-50 38.0 mAP). Surprisingly, on the more challenging OVIS, IAI achieves SOTA performance (20.6 mAP). Code is available at \url{https://github.com/zfonemore/IAI}.
\end{abstract}


\section{Introduction}
\label{sec:intro}

Instance Segmentation~\cite{lin2015microsoft,he_mask,hybrid,bmask,chen2020blendmask,zhang2020MEInst,yolact-plus-tpami2020} is an important problem in the computer vision community, which aims to detect and segment objects of specific classes in an image. Benefiting from the rapid growth of deep learning techniques, this problem has achieved great progress in recent years. Most recently, Video Instance Segmentation (VIS)~\cite{Yang_2019_ICCV}, an advanced version of instance segmentation that aims to simultaneously detect, segment, and track different objects of specific categories in videos, is attracting the attention of many researchers due to its wide application prospects, such as augmented reality and video editing.

\begin{figure}[t]
\subfigure[IAI framework]{
\begin{minipage}[!t]{0.5\textwidth}
\centering
\includegraphics[width=\textwidth]{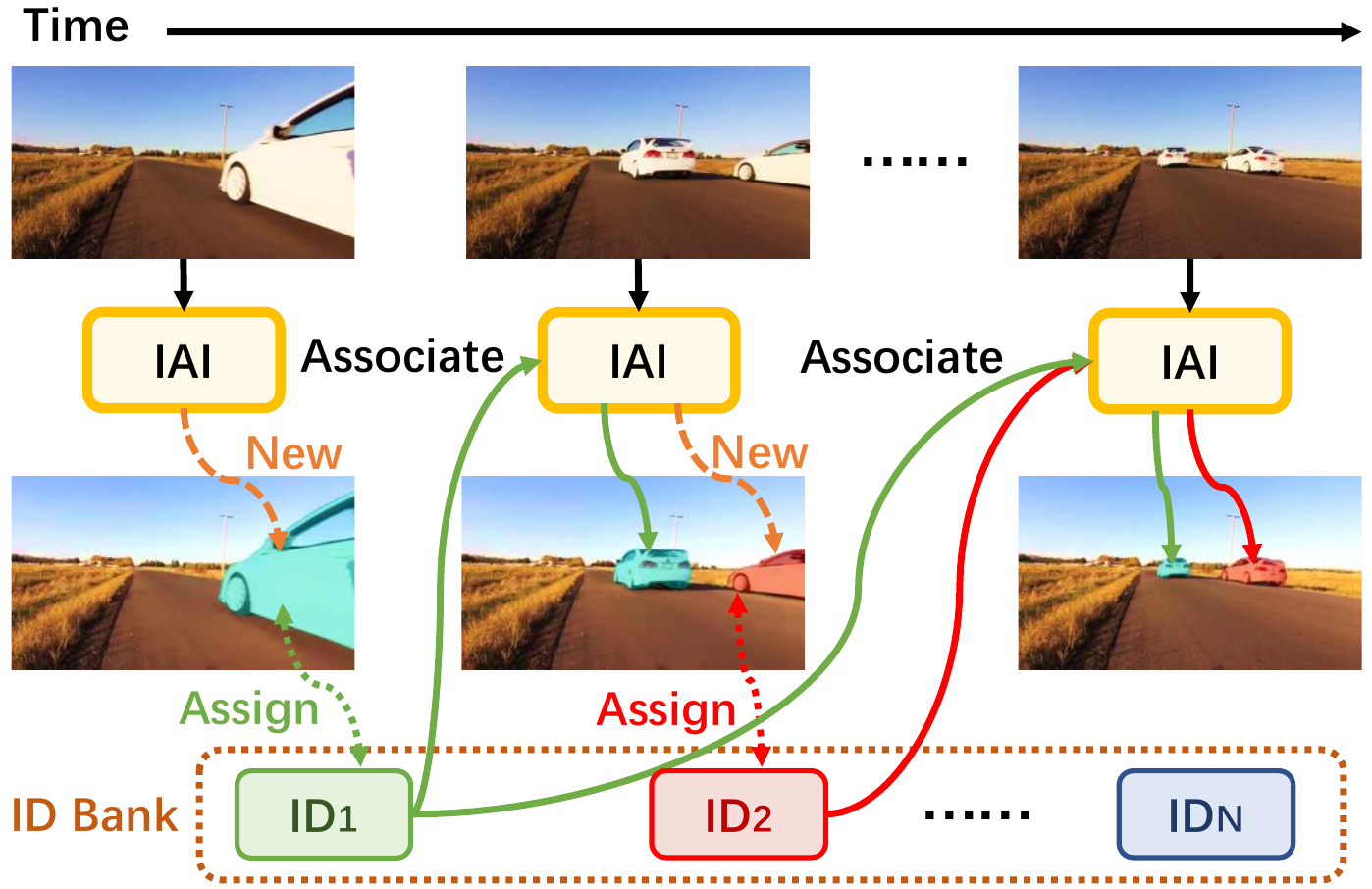}
\end{minipage}
\label{fig:itp1}
}
\subfigure[Occluded scene]{
\begin{minipage}[!t]{0.5\textwidth}
\centering
\includegraphics[width=\textwidth]{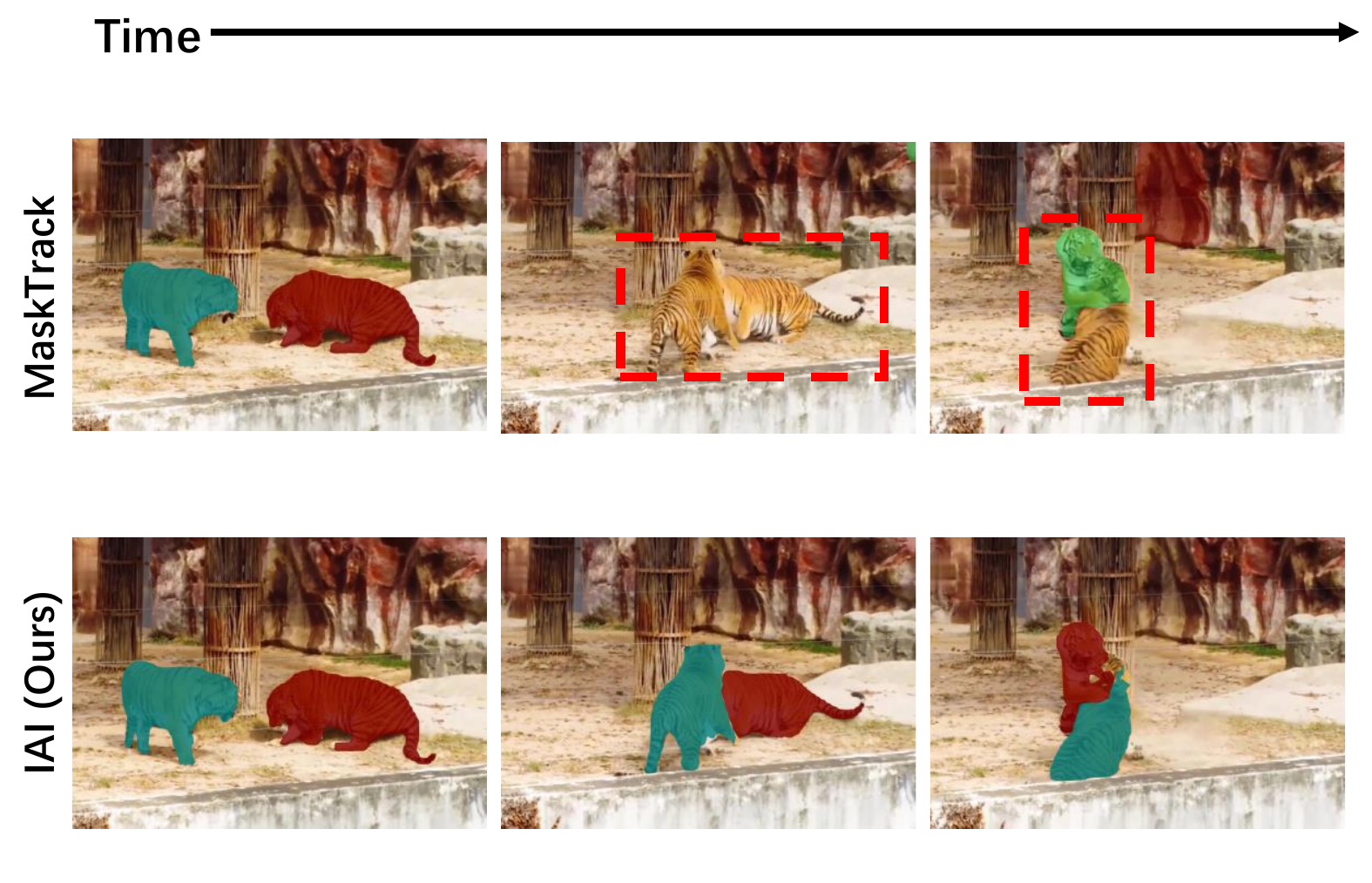}
\end{minipage}
\label{fig:occlusion}
}
\caption{(a) The illustration of the IAI paradigm. In the video, the IAI first detects a car in the initial frame and assigns it ID1. In the second frame, IAI associates the first car with ID1, and recognizes the second car as a new instance. IAI assigns ID2 to the new car. In the next frames, IAI associates these two cars using ID1 and ID2.  (b) Comparison of MaskTrack and IAI on occluded scenes from OVIS dataset.}
\end{figure}

The main challenge of VIS lies in how to assemble instance segmentation and tracking into a unified framework. Some latest progresses, e.g., MaskProp~\cite{maskprop} and VisTR~\cite{vistr}, achieved this target by handling a clip. However, these clip-based approaches fail to perform online inference and thus cannot be applied in real-world applications that requires real-time processing.
To address this issue, some online solutions~\cite{Yang_2019_ICCV,Cao_SipMask_ECCV_2020,STMask-CVPR2021,crossvis} following a tracking-by-detection paradigm are also proposed recently. Although such a paradigm can adopt temporal features for tracking, it does not utilize prior object information (e.g. appearance information and position information) to detect the corresponding ones in following frames. In view of this, we aim to design a novel online pipeline to fully exploit temporal object information and encode it for both detection and tracking processes.

To this end, we propose a new solution named Instance As Identity (IAI), as shown in Fig.~\ref{fig:itp1}. Within IAI, we first detect instances in the initial frame and assign IDentities (IDs) to them. Then, in the next frame, we directly predict IDs of instances. For those instances that fail to match any previous instances, we assign new unique IDs to them. By conducting this process on each frame, all the instances can be smoothly detected and tracked in an online manner.

To be specific, our IAI is achieved by a novel identification module and an efficient multi-object association module. The identification module consists of a new designed identification head and an identification bank. Particularly, the identification head can dynamically detect new instances and assign IDs, and the identification bank encodes the IDs and masks of objects into ID embeddings for propagating across frames. To construct an efficient multi-object association module, we propose an effective Hybrid Association Block (HAB), which adopts transformer and memory to propagate features for tracking and utilizes a classification projector to encode backbone features. It should be noted that our HAB is very different from the association module proposed in~\cite{aot}, which only works for the class-agnostic scenarios that the mask of first frame should be correctly provided by human and does not meet the requirements of VIS (i.e., one VIS model should be equipped with the ability of automatically performing instance segmentation and classification by itself). Through these two modules (i.e., identification and multi-object association), our IAI successfully achieves multiple object association at once for both detection and tracking.

To the best of our knowledge, IAI is the first VIS paradigm to use ID to unify detection and tracking in an online way. Besides, our IAI is pretty flexible and could be easily integrated with existing image segmentation models. Surprisingly, as shown in Fig.~\ref{fig:occlusion}, we find that IAI shows a strong capability on handling object occlusion, which is a key problem in VIS. We 
attribute this robustness to the design of our identification module and association module. First, because our identification module encodes multiple object information into one identification embedding, the enriched surrounding information of each object helps model to separate different instance on occlude scenes. Second, the global memory in our association module helps the model to acquire object information which is absent in long-term occlusion. 

We conduct extensive experiments on three challenging VIS benchmarks, i.e., YouTube-VIS-2019, YouTube-VIS-2021 and OVIS, to evaluate the effectiveness and generality of the proposed IAI paradigm. With ResNet-50~\cite{Resnet} as the backbone, the IAI paradigm achieves superior performance on the validation sets of YouTube-VIS-2019 (39.9 mAP) and YouTube-VIS-2021 (38.0 mAP), outperforming all the online model competitors. Particularly, IAI is the first \emph{online} method to achieve an over 40 mAP, i.e., 43.7 mAP, with ResNet-101 as the backbone on YouTube-VIS-2019. Moreover, on the more challenging OVIS dataset, our method outperforms SOTA VIS methods by a large margin (+5.2 mAP), which further proves the robustness of IAI on the occluded scenes.  

Overall, we summarize our contributions as follows:

\begin{itemize} 
    \item We propose a generic paradigm for VIS named IAI. IAI achieves superior performance on VIS benchmarks and outperforms all the online methods. 
    \item We propose a novel identification module that can re-identify the previous instance and recognize a new instance, which is the first time in VIS to track instances explicitly using IDs. 
    \item We propose a new hybrid association block as our association module, which combines backbone features with memory ID embeddings.
    
\end{itemize}  

\section{Related Work}

VIS is highly related to several tasks, such as image instance segmentation and semi-supervised video object segmentation. In this section, we provide a brief overview of recent studies in VIS and related fields.   

\textbf{Image Instance Segmentation} Image instance segmentation algorithms are mainly built on either two-stage frameworks or one-stage frameworks. Though rapid progress has been witnessed in instance segmentation, the classical two-stage architecture Mask R-CNN~\cite{he_mask} is still the most popular framework to date. Many state-of-the-art works are extended on the basis of Mask R-CNN. Mask R-CNN first predicts bounding-box proposals through a regional proposal network and then produces instance masks using the cropped features for proposals. As for one-stage algorithms, CondInst~\cite{condinst} is a good representative, which outperforms many state-of-the-art instance segmentation algorithms. CondInst adopts a dynamic instance-wise mask head to produce instance masks, thus avoiding ROI operations and enabling mask prediction with higher resolution features.    

\textbf{Semi-supervised Video Object Segmentation} Semi-supervised video object segmentation (VOS)~\cite{Perazzi2016,Pont-Tuset_arXiv_2017} targets at segmenting the given objects with the annotated object masks of the first frame in a video. Many semi-supervised VOS approaches rely on fine-tuning the first frame at test time. Some recent works~\cite{chen2018blazingly,feelvos,osmn} propose methods without fine-tuning to achieve a better run-time. STM~\cite{Oh_2019_ICCV} leverages a memory network to perform long-term propagation. CFBI~\cite{yang2020CFBI,yang2020CFBIP} utilizes the feature embedding from the target foreground object and its corresponding background collaboratively. AOT~\cite{aot} proposes a novel identification mechanism for multi-object association and utilizes a Long Short-Term Transformer to propagate information from memory frames. 

\textbf{Video Instance Segmentation} The VIS task consists of classification, segmentation, and tracking of instances in a video. Along with the YouTube-VIS 2019 datasets, \cite{Yang_2019_ICCV} proposes a representative algorithm MaskTrack R-CNN. MaskTrack R-CNN employs a tracking branch to the Mask R-CNN framework in order to link the same instance over frames. The VIS task was formally proposed in ~\cite{Yang_2019_ICCV}, and most VIS methods follow the tacking-by-detection paradigm of MaskTrack R-CNN. SipMask-VIS~\cite{Cao_SipMask_ECCV_2020} adopts a tracking branch similar to the one-stage FCOS~\cite{tian2019fcos} and YOLACT~\cite{yolact-iccv2019} framework. CompFeat~\cite{fu2021compfeat} proposes a temporal attention module and a spatial attention module to extract contextual information in temporal and spatial dimensions. STMask~\cite{STMask-CVPR2021} refines spatial features by aligning features between anchors and ground-truth bounding boxes, and designs a temporal fusion module to learn cross-frame information. CrossVIS~\cite{crossvis} is built on CondInst~\cite{condinst} and exchanges dynamic filters in two different frames to learn a more robust video-based instance representation. Although these methods are online algorithms, they are not the optimal solution since detection and tracking are conducted in two independent steps. 

Different from the tracking-by-detection paradigm, the state-of-the-art algorithm MaskProp~\cite{Bertasius_2020_CVPR} designs a mask propagation mechanism to perform detection and tracking simulatenously. MaskProp utilizes deformable convolution~\cite{deform} and attention to propagate instance features across frames. VisTR~\cite{vistr} and IFC~\cite{IFC} take advantage of the superior sequence modeling of transformers, and extend the transformer-based detection model DETR~\cite{detr} to solve VIS problem. Despite the promising performance of these methods, they are both offline methods. It remains a challenge to combine detection and tracking in an online paradigm.


\section{Method}
Given an input video $G\in R^{T \times 3 \times H \times W}$ comprising of $T$ frames of spatial size $H \times W$ , VIS task requires our method to detect, segment and track instances of a predefined category set $\omega = \{1,...,P\}$ in video $G$. To be specific, our model predicts an instance mask track $M_{G}^{i} \in \{0, 1\}^{T\times H\times W}$ with a class label $c^{i} \in \{1,...,P\}$ and a confidence score $s^{i} \in \left[ 0, 1 \right] $ for each detected instance $i$ in $G$.

\begin{figure*}[t!]
\centering
\includegraphics[width=0.85\linewidth]{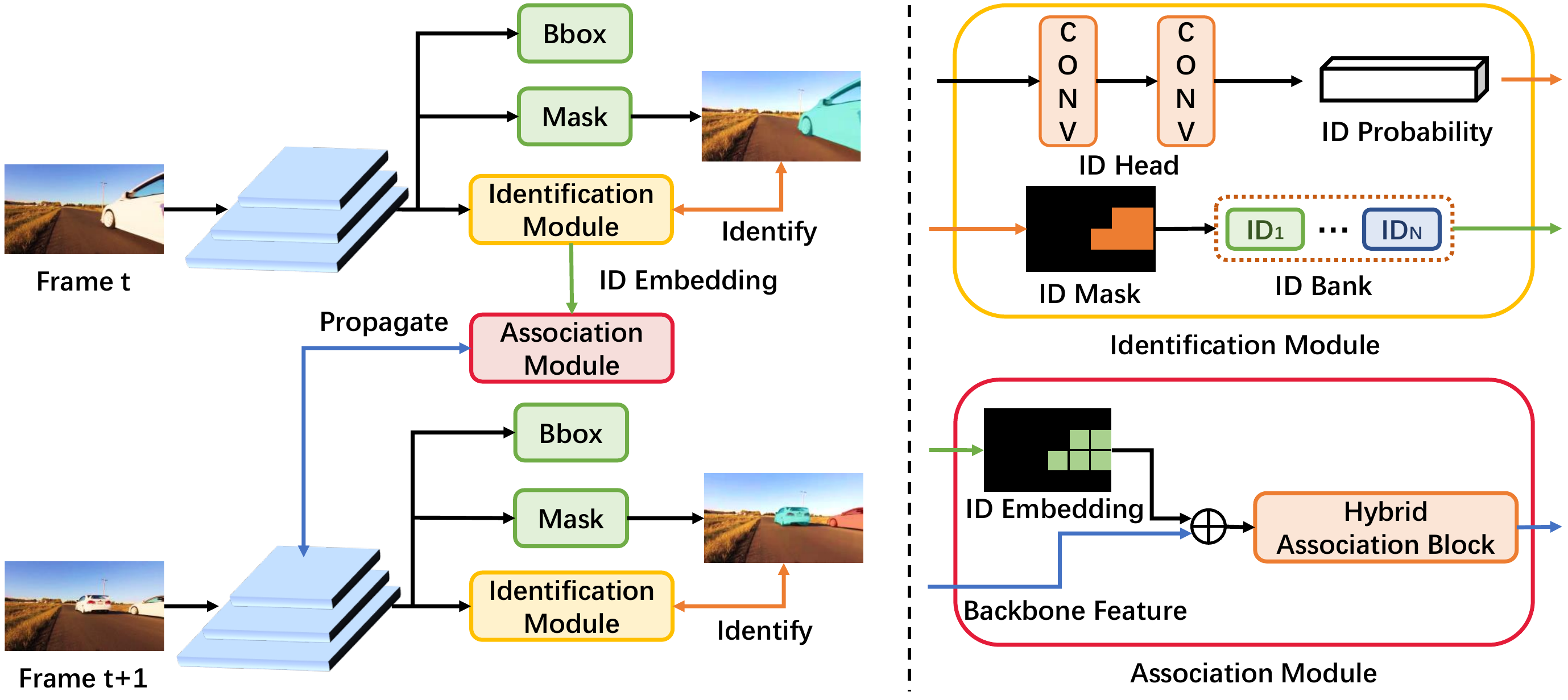}
\caption{\textbf{Overview of the IAI paradigm.} In IAI paradigm, we use identification module and association module for tracking and modeling temporal information. The identification consists of an ID head and an ID bank, the former is used to predict ID probability and the latter is used to encode ID mask into ID embedding. The association module is comprised of one hybrid association block and  used to propagate information from previous frames to frame t+1. }
\label{fig:itp2}
\end{figure*}
In order to solve this challenging problem, we propose a new online paradigm named Instance as Identity (IAI). The detailed framework of IAI is illustrated in Fig.~\ref{fig:itp2}. The IAI paradigm designs two modules to extend the original image instance segmentation framework, \emph{i.e.}, identification module and association module. In this section, we first offer a brief introduction to the basic image instance segmentation framework. Then we describe how the ID and association module are designed to combine detection and tracking in an online way.    

\subsection{Image Instance Segmentation} 
Commonly, there are two kinds of image instance segmentation frameworks: two-stage framework and one-stage framework. Since our IAI paradigm is not designed on specific image segmentation framework, we take a simplified image segmentation framework for convenience.  As it is presented in Fig.~\ref{fig:itp2}, the simplified image instance segmentation model contains backbone, bounding box head and mask head. For the image segmentation task, the model firstly uses the backbone to extract features from the image. Then the bounding box head utilizes the object features to classify and regress the bounding box. The mask head utilizes the object features to predict the mask for each instance.    

\subsection{Identification Module} 
\begin{wrapfigure}{r}{0.5\textwidth}
\centering
\includegraphics[width=0.5\textwidth]{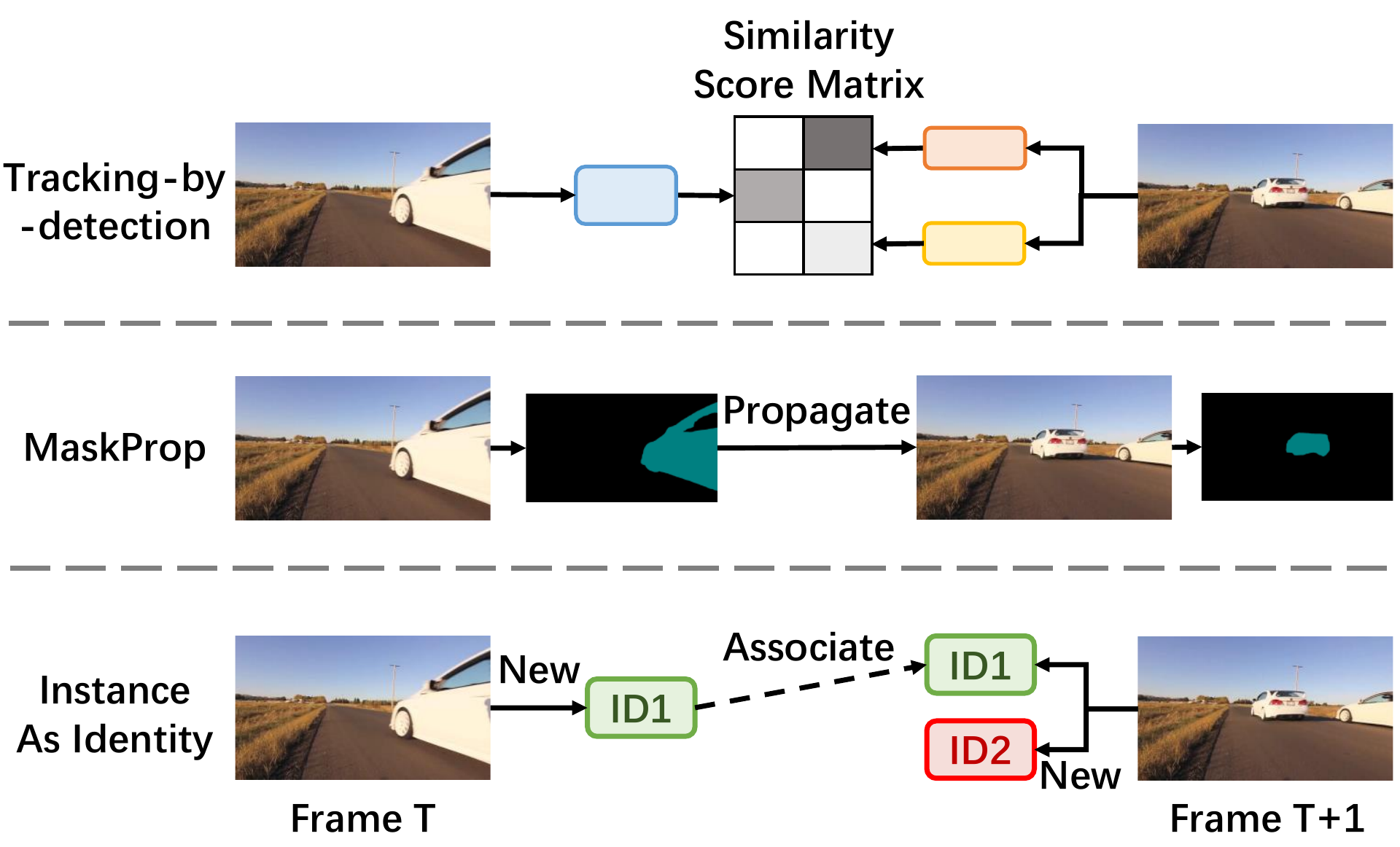}
\caption{Tracking patterns of different VIS paradigms. }
\label{fig:track}
\end{wrapfigure}

As shown in Fig.~\ref{fig:track}, previous tracking-by-detection VIS algorithms always add a tracking branch to the image instance segmentation model to achieve instance association. In this way, temporal information is only utilized for tracking but not for detection. To overcome this disadvantage, MaskProp proposes a mask propagation paradigm to combine detection and tracking. Maskprop processes each instance independently for association across and aggregates all the single-object predictions into a multi-object prediction. However, this post ensemble paradigm is not efficient for multiple object association in video tasks.

\noindent\textbf{Revisit ID Mechanism.} As for multi-object learning and understanding, ID Mechanism~\cite{aot} was recently proposed for associating and re-identifying multiple given objects in video. The ID mechanism consists of an ID embedding and an ID decoding. The ID embedding module utilizes an identity bank and a random permutation matrix to embed the mask of multiple different targets for propagation. The ID decoding module predicts the targets' probabilities using the aggregated feature. Although this ID mechanism provides a good idea for multiple object association in the video, it is impractical to directly apply it to video instance segmentation. There are two main challenges for application in VIS: (1) No targets and ground truth will be given in the first frame, which means nothing to be encoded in the ID embedding; (2) Once a new instance appears in the intermediate frame, the ID decoding module is unable to recognize it and always treats it as background.          
\noindent\textbf{Improved ID Mechanism in VIS.} To address these challenges, we propose an improved ID mechanism for VIS. In the improved ID mechanism, we will assign each instance a unique ID for the entire video. First, we will detect new instances and assign them unique IDs. Then we will use a similar ID embedding method to encode the mask of different objects. Finally, we will predict IDs for previously detected instances and recognize new instances in subsequent frames. Through this improved ID mechanism, our IAI paradigm could achieve multiple object association more effectively. 

Here, we take an example to illustrate our improved ID mechanism. In the first frame, we detect a new instances $i$ and assign it a new ID $d$. Then we encode ID and mask information of instance $i$ into ID embedding and save it to memory. In subsequent frames, the ID probability of detected instances will be predicted. With the predicted IDs, different objects are tracked across frames.

\noindent\textbf{ID Head.} In order to recognize new instances and match previous instances through ID, we design a new ID head to predict the ID probability for all object proposals. As seen in Fig.~\ref{fig:itp2}, the ID head is parallel with other heads,\emph{i.e.}, classification head, bounding box head, and mask head, and shares the same features with them, which means no additional cues are required. The ID head predicts ID probability for all the object proposals. As the number of instances could be various in different frames, we set a number N  which is large than the maximum amount of objects in a video of the benchmark (e.g. 20 for YouTube-VIS 2019) as the number of IDs in the ID head. Moreover, the ID head predicts a specific N-1-th ID for all the new instances and then assign specific IDs for them. We use IDs from 0 to N-2 to denote the detected instances in previous frames, and the N-th ID means the background class. The ID head does not need elaborate design, and it employs nearly the same structure as the classification head, $e.g.$ two convolution layers in Fig.~\ref{fig:itp2}.  

\noindent\textbf{Post Processing for Inference.} As we directly predict the IDs for instances and treat each detection as a unique instance, we use a class-agnostic NMS instead of multiclass NMS. Besides, we use an average of ID score and classification score for NMS. Different from the category prediction, the ID prediction in each frame has to be unique since there could not be two same instances in one frame. The simple ID head is unable to guarantee the uniqueness of ID predictions. Thus we adopt the Hungarian algorithm~\cite{kuhn1955hungarian} to assign the unique ID with predicted ID probability as the matching cost. Since there will be various new objects through the video, we set a previously detected object number S to assign ID for new objects. If the object ID is predicted as ID N-1, we will assign it a new ID S+1 and increase S by 1 accordingly. Once S equals $N-1$, we assume there could not be more new instances in the video, and discard the newly detected instances in following frames. 

\noindent\textbf{ID Embedding.} Assume there are $L$ detected objects in current frame, after the unique ID prediction $U \in \{0,1,...,N\}^{L} $ and mask prediction $M \in \{0, 1\}^{L \times HW}$ are obtained, we produce an ID embedding to propagate these information to following frames. In order to encode ID information and mask information of multiple instances together, we combine $U$ and $M$ to generate the one hot ID masks $Y \in \{0,1\}^{ (N+1) \times HW}$,   

\begin{equation}
    Y_{U_{i}} = M_{i}, \qquad  1 \leq i \leq L.
\end{equation}

We employ a similar ID embedding method in AOT~\cite{aot} to encode the ID masks Y. In AOT, an identity bank $D \in R^{(N+1) \times C}$ with C channel dimensions is used to project different instance features into the same feature space. The ID embedding $E \in R^{HW \times C}$ is generated by,
\begin{equation}
    E = Y^{T}D.
\end{equation}

\subsection{Association Module}
\noindent\textbf{Revisit Previous VIS Methods.}  Previous tracking-by-detection methods do not propagate information of one frame to the next frame, and they store features of previous instances for tracking instead. To combine detection and tracking in a unified model, Maskprop utilizes an attention mechanism to propagate object information. However, this attention mechanism is not efficient since it requires propagating features of every instance in a frame independently, which generates many redundant computations. Moreover, MaskProp separates the video into densely overlapped clips and propagate features from the center frame to all other frames in the clip. Although this propagation manner could avoid information loss during long-term propagation, it takes tremendous computation and memory consumption to perform attention between numerous frames. 

\begin{wrapfigure}{r}{0.5\textwidth}
\centering
\includegraphics[width=0.5\textwidth]{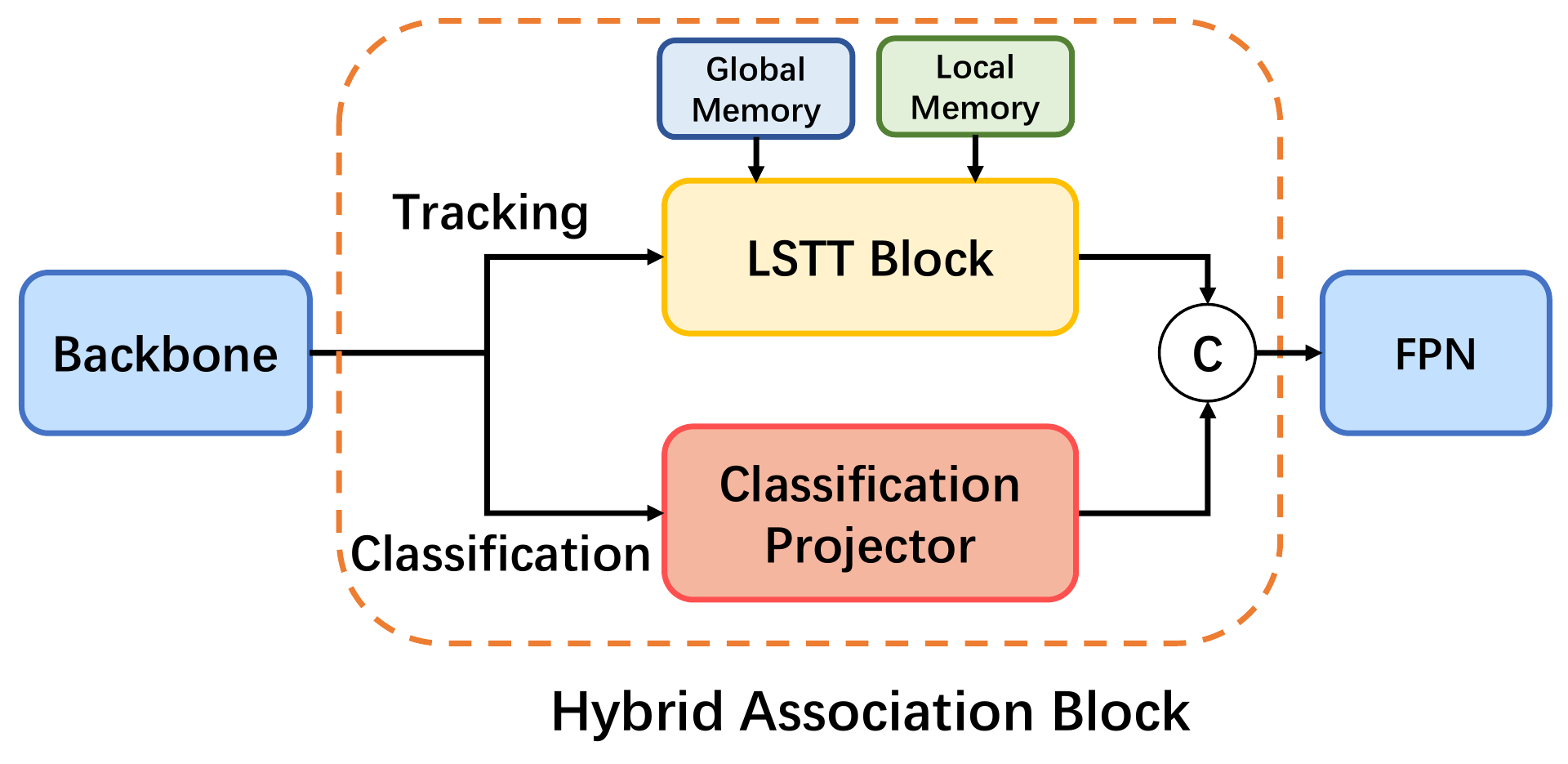}
\caption{Illustration of Hybrid Association Block.}
\label{fig:hlstt}
\end{wrapfigure}

\noindent\textbf{New Association Mechanism For VIS.} With the compact ID embedding of previous objects, we propose a new association mechanism for VIS. In our new mechanism, we utilize a local memory to store object information of last frame, and maintains a global memory of initial frame to build long-term correlation. Based on these two memory, we use attention operation to get features that contains previous object information and current information. Moreover, we adopt a parallel classification branch to encode backbone features for further classification. This new association mechanism could propagate multiple object information from previous frames to current frame at once, which is much more efficient than MaskProp.

\noindent\textbf{Hybrid Association Block (HAB).} To implement this new association methcanism, as shown in Fig.~\ref{fig:itp2}, we propose an HAB, which is extended on the LSTT block in previous VOS method AOT~\cite{aot}. As shown in Fig.~\ref{fig:hlstt}, the new HAB contains a classification projector for additional classification in VIS task. In detail, the LSTT block conducts attention between backbone features and ID embeddings from global and local memory, which learns correlation between frames for object tracking. As for the classification projector, it is 1$\times$1 convolution to encode backbone features for classification. For the output of HAB, we concat outputs from two branches to form the final output.

\subsection{Training Details}
As for training, we follow the sequential training strategy in \cite{yang2020CFBI}, in which 5 frames in a video are randomly sampled to form a sequence. For each sequence, we first assign IDs for instances in the sequence since there are no ground truth IDs in the original YouTube-VIS dataset. We assign IDs for instances per frame from 0 to N-1, \emph{e.g.} the first instance assigned 0, the second assigned 1 and so on. One important case should be mentioned is in the frame one instance first appears, the ground truth ID of it should be assigned N. 

We train detection and tracking in an end-to-end way, and the loss is 

\begin{equation}
    L = L_{cls} + L_{bbox} + L_{mask} + L_{id},
\end{equation}
where $L_{cls}$, $L_{bbox}$ and $L_{mask}$ represent the classification loss, bounding box loss and mask loss in image instance segmentation model~\cite{he_mask,condinst}. $L_{id}$ denotes the ID loss, which is implemented with a similar function like $L_{cls}$. For example, we use focal loss~\cite{focal} for ID loss when combining with CondInst,
\begin{equation}
    L_{id} = -\alpha_{t}(1-p_{i}(d))^{\lambda}log(p_{i}(d)),
\end{equation}
where $p_{i}(d)$ is the probability of assigning ID $d$ to instance $i$, $\alpha_{t}$ and $\lambda$ follow the definition in \cite{focal}.

\section{Experiment}

In this section, we conduct extensive experiments to evaluate IAI on three VIS benchmarks, YouTube-VIS-2019~\cite{Yang_2019_ICCV}, YouTube-VIS-2021~\cite{YouTube-2021} and OVIS~\cite{ovis}. 
\begin{itemize}
    \item \textbf{YouTube-VIS-2019} is the first large-scale benchmark to video instance segmentation, which consists of 2,883 high-resolution YouTube videos. The dataset is annotated with 4883 unique objects from 40 common categories and contains about 131k instance masks. 
    \item \textbf{YouTube-VIS-2021} is extended on the basis of the YouTube-VIS-2019, with more videos and a modified category label set. The YouTube-VIS-2021 contains 3,859 high-resolution YouTube videos, 8,171 unique video instances and approximately 232k high-quality manual annotated masks.
    \item \textbf{OVIS} is a large-scale benchmark for video instance segmentation, which aims to perceive object occlusion in videos. The OVIS dataset consists of 901 videos with severe object occlusions. The dataset is annotated with 5,223 unique instances from 25 commonly seen categories.
\end{itemize}

We use the average precision (mAP) and average recall (AR) defined in \cite{Yang_2019_ICCV} as the evaluation metric. Following previous works, we evaluate all results on validation set through official evaluation servers. 

\subsection{Implementation Details}
\noindent\textbf{Settings.} The model is implemented with MMDetection-2.11~\cite{mmdetection}. For training, we initialize our model with weights of corresponding instance segmentation model pre-trained on COCO train2017. The instance segmentation models are pretrained with 12 epochs. The VIS network is optimized with AdamW optimizer setting the initial learning rate to 10$^{-4}$, weight decay to 10$^{-4}$. The learning rate is reduced by a factor of 10 at 9 and 12 epochs. Specifically, the backbone's learning rate is set to 0.1 of the network's, and weight decay is set to 0.9 of the network's to avoid overfitting. The input size of each frame is resized to 360 $\times$ 640. The number of IDs in ID head is set to 20.  

We randomly sample 5 frames from a video to form a sequence for training. We train our model on VIS datasets with 1$\times$ schedule, $i.e.$, 12 epochs. The models are trained on 4 Tesla V100 GPUs with batch size of 16. During inference, the video is processed frame by frame without any test time augmentation. The FPS data of inference is measured on NVIDIA RTX A6000. 

\begin{table*}[t]
\caption{Comparisons with SOTA VIS methods on \textbf{YouTube-VIS-2019} val set. ``$\checkmark$'' under ``Aug." means multi-scale augmentation for training, ``\checkmark\checkmark" indicates stronger augmentation or additional data. ``Temp.'' means modeling temporal information for detection. We follow \cite{lin2019tsm} to define whether the VIS algorithm is online or offline.}
   \centering
   \begin{tabular}{l|c|c|c|ccccc|c}
   \noalign{\hrule height 1pt}
   Methods   & Backbone & Aug. & Temp. & mAP & AP$_{50}$ & AP$_{75}$ & AR$_{1}$ & AR$_{10}$ & FPS \\
   \hline
   \multicolumn{9}{c}{\textbf{offline methods}} \\
    \hline
    \rowcolor{gray!20} & ResNet-50 & & &30.6 & 50.7 & 33.5& 31.6 &37.1 &4.4 \\
    \rowcolor{gray!20} \multirow{-2}*{STEm-Seg \cite{stemseg}} & ResNet-101 &\multirow{-2}*{\checkmark \checkmark} & \multirow{-2}*{\checkmark} &34.6 &55.8 &37.9 &34.4& 41.6& 2.1 \\
    \multirow{2}*{MaskProp \cite{maskprop}}  & ResNet-50 &  \multirow{2}*{\checkmark \checkmark} & \multirow{2}*{\checkmark} & 40.0 & - & 42.9 & - & - & $<$6.2 \\
     & ResNet-101 &  & &42.5 & - & - &45.6 &- & $<$5.6 \\
    \rowcolor{gray!20} & ResNet-50 & & &36.2 &59.8 &36.9 &37.2 &42.4 & 30.0 \\
    \rowcolor{gray!20} \multirow{-2}*{VisTR \cite{vistr}}& ResNet-101 & \multirow{-2}*{\checkmark} & \multirow{-2}*{\checkmark} &40.1 &64.0 &45.0 &38.3 &44.9 & 27.7 \\
    \multirow{2}*{Seq Mask R-CNN \cite{seq}} & ResNet-50 & \multirow{2}*{\checkmark\checkmark} & \multirow{2}*{\checkmark} &40.4 & 63.0 & 43.8 & 41.1 & 49.7 & - \\
    & ResNet-101 & & &\textbf{43.8} & \textbf{65.5} & \textbf{47.4} & \textbf{43.0} & \textbf{53.2} & - \\
    \rowcolor{gray!20} & ResNet-50 & & & \textbf{41.2} & \textbf{65.1} & \textbf{44.6} & \textbf{42.3} & \textbf{49.6} & \textbf{107.1} \\
    \rowcolor{gray!20} \multirow{-2}*{IFC \cite{IFC}}& ResNet-101 & \multirow{-2}*{\checkmark} & \multirow{-2}*{\checkmark} &42.6 &66.6 &46.3 &43.5 &51.4 & 89.4 \\
    \hline
    \multicolumn{9}{c}{\textbf{online methods}} \\
    \hline
    \rowcolor{gray!20}  & ResNet-50 & & & 30.3 & 51.1 & 32.6 & 31.0 & 35.5 & 32.8 \\
    \rowcolor{gray!20} \multirow{-2}*{MaskTrack R-CNN \cite{Yang_2019_ICCV}} & ResNet-101 & & &31.9 &53.7 &32.3 &32.5 &37.7 & 28.6 \\
     
    SipMask-VIS \cite{Cao_SipMask_ECCV_2020} & ResNet-50 &  \checkmark & &33.7 & 54.1 &35.8 &35.4 &40.1 &34.1 \\ 
     
    \rowcolor{gray!20}  & ResNet-50 & & &33.5 &52.1 &36.9 &31.1 &39.2 & 28.6 \\
    \rowcolor{gray!20} \multirow{-2}*{STMask \cite{STMask-CVPR2021}}& ResNet-101& \checkmark &&36.8 &56.8 &38.0 &34.8 &41.8& 23.4 \\
    QueryInst-VIS \cite{queryinst_2021_ICCV,QueryTrack}  
    & ResNet-50& \checkmark & &36.2 &56.7 &39.7 &36.1 &42.9& 32.3 \\
     \rowcolor{gray!20} \multirow{1}*{CompFeat \cite{fu2021compfeat}}  & ResNet-50 & \checkmark \checkmark & \checkmark&35.3 &56.0 &38.6 &33.1 & 40.3 & - \\
      & ResNet-50 & & &34.8 & 56.1 &36.8& 35.8& 40.8 & 23.0 \\
     \multirow{-2}*{SG-Net \cite{sgnet}}& ResNet-101 & \multirow{-2}*{\checkmark} & &36.3 & 57.1 &39.6 &35.9 &43.0 & 19.8 \\
    \rowcolor{gray!20 }  & ResNet-50 & \checkmark & &36.3 &56.8 &38.9 &35.6 & 40.7 & 39.8 \\
     \rowcolor{gray!20 }\multirow{-2}*{CrossVIS \cite{crossvis}} & ResNet-101 & & &36.6 &57.3 &39.7 &36.0 &42.0 & 35.6 \\
    
     & ResNet-50 & & &37.9 &58.8 & 42.1 &38.7 & 46.8 & 32.8 \\
     & ResNet-50$*$ & \checkmark &\multirow{-2}*{\checkmark} &\textbf{39.9} &\textbf{62.3} & \textbf{43.9} &\textbf{40.1} & \textbf{46.8} & 32.8 \\
     &  ResNet-101 & & & 41.0 & 61.3 & 45.3 & 40.8 & 47.5 & 28.4 \\
    \multirow{-4}*{\textbf{IAI+CondInst}}& ResNet-101$*$ & \checkmark &\multirow{-2}*{\checkmark} & \textbf{43.7} & \textbf{67.2} & \textbf{48.4} & \textbf{41.7} & \textbf{50.0} & 28.4 \\
   \noalign{\hrule height 1pt}
   \end{tabular}
$*$ Accuracy and speed becomes better because we fix some bugs in inference.
\label{tab:vis_2019}
\end{table*}

\begin{table}[t]
\caption{Comparisons with SOTA VIS methods on \textbf{YouTube-VIS-2021} val set.} 
   \setlength{\tabcolsep}{6pt}
   \begin{tabular*}{\hsize}{l|c|c|ccccc}
   \noalign{\hrule height 1pt}
   Methods & Backbone & Aug. & mAP & AP$_{50}$ & AP$_{75}$ & AR$_{1}$ & AR$_{10}$ \\
   \hline
   \hline
    MaskTrack R-CNN & \multirow{6}*{ResNet-50}& &  28.6 &48.9 &29.6 &26.5 &33.8 \\
    SipMask-VIS  && \checkmark &31.7 & 52.5 &34.0 &30.8 &37.8  \\
    CrossVIS  &&  &33.3 &53.8 &37.0 &30.1 &37.6 \\
    CrossVIS  && \checkmark &34.2 & 54.4 &37.9 &30.4 &38.2  \\
    IFC  && \checkmark &35.2 & 57.2 &37.5 & - & -  \\
    \textbf{IAI+CondInst} &&  &\textbf{38.0} &\textbf{59.1} &\textbf{43.0} &\textbf{34.8} & \textbf{44.5} \\
   \noalign{\hrule height 1pt}
   \end{tabular*}
\label{tab:vis_2021}
\end{table}
\begin{table}[t]
\caption{Comparisons with SOTA VIS methods on \textbf{OVIS} val set.} 
   \setlength{\tabcolsep}{6pt}
   \begin{tabular*}{\hsize}{l|c|c|ccccc}
   \noalign{\hrule height 1pt}
   Methods & Backbone & Aug. & mAP & AP$_{50}$ & AP$_{75}$ & AR$_{1}$ & AR$_{10}$ \\
   \hline
   \hline
    MaskTrack R-CNN & \multirow{8}*{ResNet-50}& &  10.8 & 25.3 & 8.5 & 7.9 & 14.9 \\
    SipMask-VIS  && \checkmark &10.2 & 24.7 & 7.8 & 7.9 & 15.8  \\
    STEm-Seg && \checkmark &13.8 & 32.1 & 11.9 & 9.1 & 20.0 \\
    QueryInst-VIS  && \checkmark &14.7 & 34.7 & 11.6 & 9.0 & 21.2 \\
    STMask  &&  &15.4 & 33.8 & 12.5 & 8.9 & 21.3 \\
    CrossVIS  && \checkmark &14.9 & 32.7 &12.1 &10.3 & 19.8 \\
    CMaskTrack R-CNN && \checkmark &15.4 & 33.9 &13.1 &9.3 &20.0  \\
    \textbf{IAI+CondInst} & &  &18.5 &36.8 &18.0 &11.7 & 24.0 \\
    \textbf{IAI+CondInst$*$} & &  &\textbf{20.6} &\textbf{38.9} &\textbf{20.3} &\textbf{11.9} & \textbf{25.8} \\
   \noalign{\hrule height 1pt}
   \end{tabular*}
   $*$ means the model is pretrained on YouTube-VIS 2019 dataset.
\label{tab:ovis}
\end{table}

\subsection{Main Results}
\noindent\textbf{YouTube-VIS-2019 Dataset.} We apply our IAI paradigm on one-stage segmentation model CondInst, and compare it with state-of-the-art methods in Tab.~\ref{tab:vis_2019}. With simple multi-scale training augmentation, our method achieves 39.9 mAP with ResNet-50 backbone, which outperforms all the online methods in Tab.~\ref{tab:vis_2019}. Moreover, our method even outperforms STEm-seg, STMask and CrossVIS with a stronger Resnet-101 backbone. With the ResNet-101 backbone, our method surpasses the recently proposed online method STMask and CrossVIS by about 7 points in mAP with simple multi-scale augmentation. As for speed, our method achieves a real-time speed at 32.8 FPS and 101 FPS with 4 GPUs. Compared with other online algorithms, we argue that utilizing prior information for detection during inference partly slows down IAI paradigm.  

As for the state-of-the-art offline method MaskProp, we argue the high performance of MaskProp partly comes from its combination with multiple strong networks, $e.g.$ Spatiotemporal Sampling Network~\cite{SSN}, Hybrid Task Cascade mask head~\cite{hybrid}, High-Resolution Mask Refinement post-process, and complex training augmentations, $e.g.$ extra OpenImages~\cite{OpenImages} datasets and longer training schedule. Meanwhile, MaskProp requires huge computation and memory cost to achieve high performance, which impedes it from online scenarios. We aim to design an efficient online paradigm for VIS, and our method can be integrated with different image segmentation models to solve VIS in an online fashion. Overall, the experimental results prove the effectiveness of the new paradigm.   

\noindent\textbf{YouTube-VIS-2021 Dataset.} YouTube-VIS-2021 dataset is an upgraded version of YouTube-VIS-2019 dataset, with more videos and an improved class set. We evaluate our method on this new dataset and compare it with some state-of-the-art approaches. As shown in Tab.~\ref{tab:vis_2021}, our algorithm outperforms SipMask-VIS and CrossVIS by a large margin without any training augmentations. The experiment results further demonstrate IAI's advantage over other paradigms. 

\noindent\textbf{OVIS Dataset.} To further prove the effectiveness and robustness of our method, we evaluate our method on the OVIS dataset. The OVIS dataset is much harder than YouTube-VIS-2019 and YouTube-VIS-2021 dataset, which contains more instances and more occluded cases per video. As shown in Tab.~\ref{tab:ovis}, our methods outperforms SOTA VIS methods by a large margin (+5.2 mAP), which indicates strong ability of our methods on dealing with object occlusion.

\begin{figure}[t]
\centering
\includegraphics[width=1\textwidth]{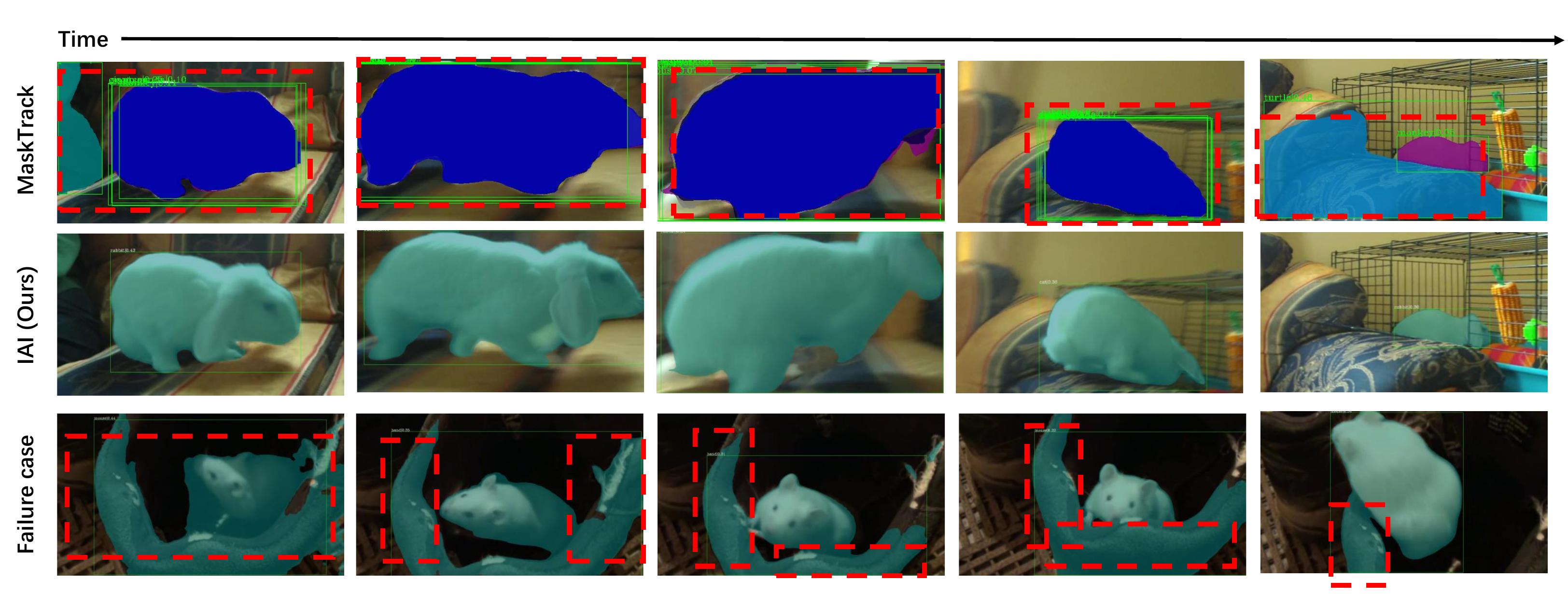}
\caption{Qualitative results. (top) Compared with MaskTrack R-CNN, IAI could make better use of temporal information, and performs well even in fast-moving scene. (bottom) Some errors are propagated once mistake happens in previous frames. }
\label{fig:visualization}
\end{figure}

\noindent\textbf{Qualitative Results.} Fig.~\ref{fig:visualization} visualizes some qualitative results of IAI in comparison with MaskTrack R-CNN. As shown in Fig.~\ref{fig:visualization}, IAI segments and tracks object more accurately than MaskTrack R-CNN, especially in fast-moving scenes. IAI makes a better fuse of temporal information, which enables it to handle motion blur and fast-moving target tracking. However, IAI relies on the segmentation quality of the first frame, once a mistake happens in the first frame, IAI might propagate it to the next frames.

\subsection{Ablation Study}
In this section, we conduct extensive ablation study experiments to prove the general effectiveness of our method. All the experiments are conducted on the YouTube-VIS 2019 dataset. All models are with ResNet-50 FPN~\cite{fpn} as backbone, and trained in 1 $\times$ schedule without any augmentation. 

\begin{table}[t]
 \begin{minipage}[!t]{0.52\textwidth}
    \vspace{-10pt}
    \caption{Experiments of the identification module and association module.}
   \setlength{\tabcolsep}{2pt}
   \begin{tabular*}{\hsize}{c|c|ccc}
   \noalign{\hrule height 1pt}
    Identification & Association  & mAP & AP$_{50}$ & AP$_{75}$ \\
   \hline
       & & 24.0 & 40.4 & 23.0 \\
      \checkmark & &12.6  & 19.5 & 13.3 \\
      & \checkmark &24.5  & 40.9 & 25.0 \\
      \checkmark & \checkmark &37.9  & 58.8 & 42.1 \\
   \noalign{\hrule height 1pt}
   \end{tabular*}
    \label{tab:associate module}
 \end{minipage}\quad
  \begin{minipage}[!t]{0.45\textwidth}
    \caption{Experiments of three key components of the HAB block.}
   \setlength{\tabcolsep}{2pt}
   \begin{tabular*}{\hsize}{c|c|c|ccc}
   \noalign{\hrule height 1pt}
    Local & Global & Class & mAP & AP$_{50}$ & AP$_{75}$ \\
   \hline
       & & & 12.6  & 19.5 & 13.3 \\
      \checkmark & & \checkmark&32.9  & 50.0 & 36.7 \\
      & \checkmark & \checkmark&34.9  & 53.1 & 38.2 \\
      \checkmark & \checkmark & &36.9  & 56.7 & 40.1 \\
      \checkmark & \checkmark & \checkmark &37.9  & 58.8 & 42.1 \\
   \noalign{\hrule height 1pt}
   \end{tabular*}
    \label{tab:memory}
 \end{minipage}
\end{table}

\noindent\textbf{Identification module and association module.} We conduct ablation study to prove the effectiveness of two key modules of our method. As shown in Tab.~\ref{tab:associate module}, the basic model without ID and association module performs poor, and with both two modules, our IAI achieves superior accuracy. Another important observation is that both two modules are necessary for our IAI paradigm. Without association module, the model could not model previous information and predict accurate IDs. Even worse, the identification module will lead to terrible performance because the ID supervision adds extra noise to model training. Without identification module, the model could not track the instances and gets similar performance to image model.

\noindent\textbf{HAB block.}  As the HAB block is the basic component of our association module, we conduct some experiments to verify the effectiveness. We study three key components of the HAB block in Tab.~\ref{tab:memory}: global memory, local memory and classification projector. From the results, we can find that both three components are effective in our IAI paradigm. The global, local memory and classification projector could bring an improvement of 5.0, 3.0 and 1.0 mAP separately.

\begin{wraptable}{r}{0.6\textwidth}
    \caption{Comparisons with other paradigms on different image instance segmentation frameworks. ``Track" means tracking-by-detection paradigm.}
    \vspace{10pt}
  \centering
   \setlength{\tabcolsep}{3pt}
   \begin{tabular*}{\hsize}{l|c|ccc}
   \noalign{\hrule height 1pt}
   Image Model  & Paradigm & mAP & AP$_{50}$ & AP$_{75}$ \\
   \hline
   \hline
    \multirow{2}{*}{Mask R-CNN} 
      & Track &  30.3  & 51.1  & 32.6\\
      & IAI & 31.7  & 49.9 & 34.6\\
      \hline
    \multirow{2}{*}{CondInst}  
      & Track &32.1 & - & -\\
      & IAI &37.9  & 58.8 & 42.1\\
   \noalign{\hrule height 1pt}
   \end{tabular*}
    \label{tab:framework}
\end{wraptable}

\noindent\textbf{Image segmentation model.} To prove the generality of IAI paradigm, we experiment with both one-stage and two-stage models. In the experiment, we choose CondInst and Mask R-CNN for the representative of one-stage and two-stage models separately. In Tab.~\ref{tab:framework}, we compare our paradigm with the tracking-by-detection paradigm on two image segmentation models. From the results, we could see that IAI paradigm outperforms the tracking-by-detection paradigm on both one-stage and two-stage segmentation models. As for why the IAI on CondInst bring a larger improvement than IAI on MaskTrack, we argue that IAI benefits more from a better image model because more accurate segmentation of first frame (no previous information) can lead to better propagation for next frames. 

\noindent\textbf{ID loss function.} As we introduce a new ID loss in IAI paradigm, we study the effect of the different ID loss functions in Tab.~\ref{tab:idloss}. From the results, we could find that focal loss~\cite{focal} brings a 1.5 mAP improvement over cross-entropy (CE) loss. As the classification loss function is focal loss, this comparison proves that keeping ID loss function consistent with the classification loss function is enough for good performance, which indicates that no additional design is required for the ID loss function.   

\noindent\textbf{ID head convolution layer number.} As the ID head plays an important role in the identification module, and we evaluate the effect of different ID head convolution layer numbers on performance. As shown in Tab.~\ref{tab:idconv}, more convolution layers do not bring obvious improvement, and using 4 convolution layers even gets worse accuracy. A possible reason is that ID information is relatively simple compared with appearance information. The appearance information might contain color, shape and other characteristics, while ID information only focuses on similarity between instances. Since ID information is easy to capture, increasing parameters is unable to boost the performance and might cause overfitting.

\begin{table}[!t]
 \begin{minipage}[!t]{0.48\textwidth}
    \vspace{-10pt}
    \caption{Experiments of different ID loss functions.}
   \setlength{\tabcolsep}{8pt}
   \begin{tabular*}{\hsize}{c|ccc}
   \noalign{\hrule height 1pt}
    $L_{id}$ & mAP & AP$_{50}$ & AP$_{75}$ \\
   \hline
   \hline
       CE loss &  36.4  & 53.8 & 40.9 \\
       Focal loss & 37.9  & 58.8 & 42.1 \\
   \noalign{\hrule height 1pt}
   \end{tabular*}
    \label{tab:idloss}
 \end{minipage}\quad\quad
 \begin{minipage}[!t]{0.46\textwidth}
    \caption{Experiments of different ID head convolution layer numbers.}
   \setlength{\tabcolsep}{8pt}
   \begin{tabular*}{\hsize}{c|ccc}
   \noalign{\hrule height 1pt}
    ID Head & mAP & AP$_{50}$ & AP$_{75}$ \\
   \hline
   \hline
       2Conv &  37.9  & 58.8 & 42.1 \\
       3Conv & 38.1  & 61.3 & 40.9 \\
       4Conv & 35.3  & 53.6 & 37.9 \\
   \noalign{\hrule height 1pt}
   \end{tabular*}
    \label{tab:idconv}
 \end{minipage}
\end{table}

\section{Conclusion}

In this paper, we introduce IAI, a novel generic online paradigm for video instance segmentation. The new IAI paradigm successfully utilizes prior object information for both detection and tracking in an online way, and perform multiple object association at once. These advantages make IAI outperform all the online video instance segmentation methods in the challenging YouTube-VIS benchmarks. Notably, the IAI paradigm shows obvious advantages over previous tracking-by-detection paradigm on occluded scenes, outperforming these methods by a large margin on OVIS benchmark.  We hope our IAI paradigm could perform as a strong baseline in the VIS and OVIS task, and contribute to future research on video understanding tasks. 

\smallskip
\noindent
\textbf{Acknowledgment.}
This work was supported in part by the National NSF of China (No.62120106009), the Fundamental Research Funds for the Central Universities (No. K22RC00010).

\par\vfill\par

\clearpage
%
%
\bibliographystyle{splncs04}
\bibliography{egbib}

\begin{thebibliography}{10}
\providecommand{\url}[1]{\texttt{#1}}
\providecommand{\urlprefix}{URL }
\providecommand{\doi}[1]{https://doi.org/#1}

\bibitem{stemseg}
Athar, A., Mahadevan, S., O{\v{s}}ep, A., Leal-Taix{\'e}, L., Leibe, B.:
  Stem-seg: Spatio-temporal embeddings for instance segmentation in videos. In:
  ECCV (2020)

\bibitem{maskprop}
Bertasius, G., Torresani, L.: Classifying, segmenting, and tracking object
  instances in video with mask propagation. In: CVPR (2020)

\bibitem{Bertasius_2020_CVPR}
Bertasius, G., Torresani, L.: Classifying, segmenting, and tracking object
  instances in video with mask propagation. In: CVPR (2020)

\bibitem{SSN}
Bertasius, G., Torresani, L., Shi, J.: Object detection in video with
  spatiotemporal sampling networks. In: ECCV (2018)

\bibitem{yolact-iccv2019}
Bolya, D., Zhou, C., Xiao, F., Lee, Y.J.: Yolact: {Real-time} instance
  segmentation. In: ICCV (2019)

\bibitem{yolact-plus-tpami2020}
Bolya, D., Zhou, C., Xiao, F., Lee, Y.J.: Yolact++: Better real-time instance
  segmentation. TPAMI  (2020)

\bibitem{Cao_SipMask_ECCV_2020}
Cao, J., Anwer, R.M., Cholakkal, H., Khan, F.S., Pang, Y., Shao, L.: Sipmask:
  Spatial information preservation for fast image and video instance
  segmentation. In: ECCV (2020)

\bibitem{detr}
Carion, N., Massa, F., Synnaeve, G., Usunier, N., Kirillov, A., Zagoruyko, S.:
  Conditional convolutions for instance segmentation. In: ECCV (2020)

\bibitem{chen2020blendmask}
Chen, H., Sun, K., Tian, Z., Shen, C., Huang, Y., Yan, Y.: {BlendMask}:
  Top-down meets bottom-up for instance segmentation. In: CVPR (2020)

\bibitem{hybrid}
Chen, K., Pang, J., Wang, J., Xiong, Y., Li, X., Sun, S., Feng, W., Liu, Z.,
  Shi, J., Ouyang, W., Loy, C.C., Lin, D.: Hybrid task cascade for instance
  segmentation. In: CVPR (2019)

\bibitem{mmdetection}
Chen, K., Wang, J., Pang, J., Cao, Y., Xiong, Y., Li, X., Sun, S., Feng, W.,
  Liu, Z., Xu, J., Zhang, Z., Cheng, D., Zhu, C., Cheng, T., Zhao, Q., Li, B.,
  Lu, X., Zhu, R., Wu, Y., Dai, J., Wang, J., Shi, J., Ouyang, W., Loy, C.C.,
  Lin, D.: {MMDetection}: Open mmlab detection toolbox and benchmark. arXiv
  preprint arXiv:1906.07155  (2019)

\bibitem{chen2018blazingly}
Chen, Y., Pont-Tuset, J., Montes, A., Van~Gool, L.: Blazingly fast video object
  segmentation with pixel-wise metric learning. In: CVPR (2018)

\bibitem{bmask}
Cheng, T., Wang, X., Huang, L., Liu, W.: Boundary-preserving mask r-cnn. In:
  ECCV (2020)

\bibitem{deform}
Dai, J., Qi, H., Xiong, Y., Li, Y., Zhang, G., Hu, H., Wei, Y.: Deformable
  convolutional networks. In: ICCV (2017)

\bibitem{queryinst_2021_ICCV}
Fang, Y., Yang, S., Wang, X., Li, Y., Fang, C., Shan, Y., Feng, B., Liu, W.:
  Instances as queries. In: ICCV (2021)

\bibitem{fu2021compfeat}
Fu, Y., Yang, L., Liu, D., Huang, T.S., Shi, H.: Compfeat: Comprehensive
  feature aggregation for video instance segmentation. In: AAAI (2021)

\bibitem{he_mask}
He, K., Gkioxari, G., Dollar, P., Girshick, R.: Mask r-cnn. In: ICCV (2017)

\bibitem{Resnet}
He, K., Zhang, X., Ren, S., Sun, J.: Deep residual learning for image
  recognition. In: CVPR (2016)

\bibitem{IFC}
Hwang, S., Heo, M., Oh, S.W., Kim, S.J.: Video instance segmentation using
  inter-frame communication transformers. arXiv preprint arXiv:2106.03299
  (2021)

\bibitem{kuhn1955hungarian}
Kuhn, H.W.: The hungarian method for the assignment problem. Naval research
  logistics quarterly  \textbf{2}(1-2),  83--97 (1955)

\bibitem{OpenImages}
Kuznetsova, A., Rom, H., Alldrin, N., Uijlings, J., Krasin, I., Pont-Tuset, J.,
  Kamali, S., Popov, S., Malloci, M., Kolesnikov, A., Duerig, T., Ferrari, V.:
  The open images dataset v4: Unified image classification, object detection,
  and visual relationship detection at scale. IJCV  (2020)

\bibitem{STMask-CVPR2021}
Li, M., Li, S., Li, L., Zhang, L.: Spatial feature calibration and temporal
  fusion for effective one-stage video instance segmentation. In: CVPR (2021)

\bibitem{seq}
Lin, H., Wu, R., Liu, S., Lu, J., Jia, J.: Video instance segmentation with a
  propose-reduce paradigm  (2021)

\bibitem{lin2019tsm}
Lin, J., Gan, C., Han, S.: Tsm: Temporal shift module for efficient video
  understanding. In: ICCV (2019)

\bibitem{fpn}
Lin, T.Y., Dollar, P., Girshick, R., He, K., Hariharan, B., Belongie, S.:
  Feature pyramid networks for object detection. In: CVPR (2017)

\bibitem{focal}
Lin, T.Y., Goyal, P., Girshick, R., He, K., Dollar, P.: Focal loss for dense
  object detection. In: ICCV (2017)

\bibitem{lin2015microsoft}
Lin, T.Y., Maire, M., Belongie, S., Bourdev, L., Girshick, R., Hays, J.,
  Perona, P., Ramanan, D., Zitnick, C.L., Dollár, P.: Microsoft coco: Common
  objects in context (2015)

\bibitem{sgnet}
Liu, D., Cui, Y., Tan, W., Chen, Y.: Sg-net: Spatial granularity network for
  one-stage video instance segmentation. In: CVPR (2021)

\bibitem{Oh_2019_ICCV}
Oh, S.W., Lee, J.Y., Xu, N., Kim, S.J.: Video object segmentation using
  space-time memory networks. In: ICCV (2019)

\bibitem{Perazzi2016}
Perazzi, F., Pont-Tuset, J., McWilliams, B., {Van Gool}, L., Gross, M.,
  Sorkine-Hornung, A.: A benchmark dataset and evaluation methodology for video
  object segmentation. In: CVPR (2016)

\bibitem{Pont-Tuset_arXiv_2017}
Pont-Tuset, J., Perazzi, F., Caelles, S., Arbel\'aez, P., Sorkine-Hornung, A.,
  {Van Gool}, L.: The 2017 davis challenge on video object segmentation.
  arXiv:1704.00675  (2017)

\bibitem{ovis}
Qi, J., Gao, Y., Hu, Y., Wang, X., Liu, X., Bai, X., Belongie, S., Yuille, A.,
  Torr, P., Bai, S.: Occluded video instance segmentation. arXiv preprint
  arXiv:2102.01558  (2021)

\bibitem{condinst}
Tian, Z., Shen, C., Chen, H.: Conditional convolutions for instance
  segmentation. In: ECCV (2020)

\bibitem{tian2019fcos}
Tian, Z., Shen, C., Chen, H., He, T.: {FCOS}: Fully convolutional one-stage
  object detection. In: ICCV (2019)

\bibitem{feelvos}
Voigtlaender, P., Chai, Y., Schroff, F., Adam, H., Leibe, B., Chen, L.C.:
  Feelvos: Fast end-to-end embedding learning for video object segmentation.
  In: CVPR (2019)

\bibitem{vistr}
Wang, Y., Xu, Z., Wang, X., Shen, C., Cheng, B., Shen, H., Xia, H.: End-to-end
  video instance segmentation with transformers. In: CVPR (2021)

\bibitem{YouTube-2021}
Xu, N., Yang, L., Yang, J., Yue, D., Fan, Y., Liang, Y., Huang., T.S.:
  Youtube-vis dataset 2021 version. https://youtube-vos.org/dataset/vis (2021)

\bibitem{Yang_2019_ICCV}
Yang, L., Fan, Y., Xu, N.: Video instance segmentation. In: ICCV (2019)

\bibitem{osmn}
Yang, L., Wang, Y., Xiong, X., Yang, J., Katsaggelos, A.K.: Efficient video
  object segmentation via network modulation. In: CVPR (2018)

\bibitem{crossvis}
Yang, S., Fang, Y., Wang, X., Li, Y., Fang, C., Shan, Y., Feng, B., Liu, W.:
  Crossover learning for fast online video instance segmentation. In: ICCV
  (2021)

\bibitem{QueryTrack}
Yang, S., Fang, Y., Wang, X., Li, Y., Shan, Y., Feng, B., Liu, W.: Tracking
  instances as queries. arXiv preprint arXiv:2106.11963  (2021)

\bibitem{yang2020CFBI}
Yang, Z., Wei, Y., Yang, Y.: Collaborative video object segmentation by
  foreground-background integration. In: ECCV (2020)

\bibitem{aot}
Yang, Z., Wei, Y., Yang, Y.: Associating objects with transformers for video
  object segmentation. In: NeurIPS (2021)

\bibitem{yang2020CFBIP}
Yang, Z., Wei, Y., Yang, Y.: Collaborative video object segmentation by
  multi-scale foreground-background integration. TPAMI  (2021)

\bibitem{zhang2020MEInst}
Zhang, R., Tian, Z., Shen, C., You, M., Yan, Y.: Mask encoding for single shot
  instance segmentation. In: CVPR (2020)

\end{thebibliography}
\end{document}